# TOWARDS A GENERAL MANY-SORTED FRAMEWORK FOR DESCRIBING CERTAIN KINDS OF LEGAL STATUTES WITH A POTENTIAL COMPUTATIONAL REALIZATION

Danny A. J. Gómez-Ramírez[1] and Egil Nordqvist[2]

## ABSTRACT

Examining a 20th-century Scandinavian legal theoretical tradition, we can extract an ontological naturalistic, a logical empiristic, and a modern idealistic rationale. We introduce the mathematical syntactic figure present in the "logical empiricism" in a contemporary mathematical logic. A new formal framework for describing explicit purchase statutes (Sweden) is gradually developed and subsequently proposed. This new framework is based on a many-sorted first-order logic (MFOL) approach, where the semantics are grounded in concrete "physical" objects and situations with a legal relevance. Specifically, we present a concrete formal syntactic translation of one of the central statutes of Swedish legislation for the purchase of immovable property. Additionally, we discuss the potential implications that a subsequent development of such formalisations would have for constructing artificial agents (e.g., software) that can be used as "co-creative" legal assistance for solving highly complex legal issues concerning the transfer of property, among others.

## I. GENERAL INTRODUCTION

The 19th and 20th centuries in the European continental and Anglo-American philosophical traditions feature shifts in the paradigmatic descriptions of knowledge production. These shifts range from given categories of perception of an empiric subject with the ability to synthesise and deductively order phenomena representing materia, with the potential to appear as objects[3]; to an empiric and cognitive subject with the ability to subsume and conceptualise the internal determinations of a subject matter, with the potential to appear in itself and to reveal itself as phenomena[4]; to approved natural scientific methods[5]; and via the mathematisation of the validity of argumentation in natural languages[6] to different combinations of logical grammars[7] and theories of supporting matter[8]. Each construction carries a specific notion of "referred" and "reference", a manner of recognition, gaining access to, and organising the referred, and a cohesive and authoritative frame and a notion of nature, appearing due to the specific requirements for access and identification[9]. The analysis of the validity of arguments[10] including the references underwent in parallel an alteration, from the lifting up of entities, letting them present certain properties or constants and relate them with each other through logical connectors, validated empirically by the application of a so-called truth-table[11]; to an expansion of the recognised properties and constants with a relational symbol with expressive power from set theory, validated through the distribution of entities in a certain given domain[12]; to the expansion of the formal syntax with a formal validation through the application of a model in consideration, i.e., a formal structure that carries the "meaning" of the purely syntactic propositions[13,14].

---

[1] Professor and General Leader of the Research and Innovation Center Parque Tech, Technical University Pascual Bravo, Medellín, Colombia. E-mail: Daj.GomezRamirez@gmail.com
[2] Research Assistant, University of Osnabrück, Osnabrück, Germany. E-mail: nordqvistegil@gmail.com

---

[3] Transcendental idealism – categories, subject, phenomena, connection, unity; see I. Kant, *Critique of Pure Reason* (London 1993), 96-97, 103-118.

[4] German idealism – subject, being, being-in-itself, being-there, being-for-other; the True, the absolute Idea, Idea, concept, science; law, phenomenal world; see G. F. W. Hegel, *Science of Logic* (New York 1989), 120, 503, 536, 842-843.

[5] The construction of universal natural scientific truth criteria inspired by the refined developed mathematical and empirical tools applied for the validation, mapping and prediction of natural phenomena (ontological/scientific naturalism), J. Ward, *Naturalism and Agnosticism*, vol. 1 (London 1899), 186; and a theory of concepts based on the dynamically evolving expressive power of mathematics capable of projecting empirically verifiable relations and identities onto an obtuse materia (philosophy of symbolic forms), E. Cassirer, *The Philosophy of Symbolic Forms. vol. 1 Language* (New Haven 1955), 45-85.

[6] Mathematical logic, W. Demopoulos, P. Clark, "The Logicism of Frege, Dedekind, and Russel", in S. Shapiro (ed.), *The Oxford Handbook of Philosophy of Mathematics and Logic* (Oxford 2005), 131-132. Logical positivism, analytical philosophy, linguistic phenomenology, structural linguistics, transformal generative grammatics, L. T. F. Gamut, *Logic, Language and Meaning*, vol. 1 (Chicago 1991), 14-15.

[7] The use of methods from different mathematical disciplines such as topology and transcendental algebra to construct logical theorems in a theory of models, E. Mendelson, *Introduction to Mathematical Logic*, 5th ed. (New York 2010), 227.

[8] Epistemological models based on an intuitive empiricism: methodological naturalism (W. V. Quine, *Ontological Relativity & Other Essays* (New York 1969), 83) and phenomenology – also referred to as representational or correlational constructions, D. Papineau, "Against representationalism (about conscious sensory experience)" (2016) 24 International Journal of Philosophical Studies 324. Non-empirically founded epistemological models: modern philosophical idealism, minimal phenomenology, liberal naturalism (J. Reynolds, "Phenomenology and naturalism: a hybrid and heretical proposal" (2016), 24 International Journal of Philosophical Studies 393), naturalistic monism - as a consequence non-representational or non-correlational constructions, D. Zahavi, "The end of what? Phenomenology vs speculative realism" (2016) 24 International Journal of Philosophical Studies 293. The distinction between a contemporary Anglo-American tradition and a continental European tradition: naturalism vs. naturalistic monism, R. Winkler, C. Botha, A. Oliver, "Phenomenology and Naturalism" (2016) 24 International Journal of Philosophical Studies 85.

[9] See the notions developed for the dissolution of a representational paradigm in favour of a non-representational paradigm, conditioning different referential activities (a mistake of direction as opposed to a mistake of the senses), Bruno Latour, *An Inquiry in to Modes of Existence* (Cambridge Mass. 2013). In a representational model: Reproduction (REP), Reference (REF), Preposition (PRE), Organisations (ORG), Nature (REP.ORG), Frame, Aggregation, Mini-transcendence, Meta-dispatcher. In a non-representational model: Attachment (ATT), Nature (ATT.ORG), Network (NET), Value (VAL), Scaling, ibid.

[10] A formal system that consists of a set of syntactical expressions, constituted by well formed formulas, a finite sequence of symbols from a given alphabet that is part of the formal language: logical constants, the logical variables, and the auxiliary signs (terms). A formal language can be identified through a set of formulas in this language. A formula is a syntactical object that can be given semantic meaning through interpretation, Gamut, *Logic, Language and Meaning*, 1, 25.

As the legal activity includes the attribution of qualities and relations to entities in a legal context, as well as the issue of the coherence and legitimacy of the legal systems, the reference-referred amalgamates, and the models for argumentative analysis offer a palette for a possible reconditioning of the legal activity, i.e., an altered relation between theory and practice[15].

In the 18th and 19th centuries, traditional legal concepts, such as ownership, originally conceived as representations of the inert ideal relations between objects predetermined in a mathematical context[16], were re-interpreted as relations between objects perceived in terms of categories of space, time, substance (or object) and causality and deductively systematised[17] or as subsumed and conceptualised representations of internal determinations of a subject matter[18]. The application of the categories subsumations and conceptualisations installed a rational subject, respectively, bound[19] and not bound[20], geographically and historically, as a reflection of the constructed legal systems.

---

[11] In propositional logic, there are logical connectors: "and" (conjunction, ∧), "or" (disjunction, ∨), "if…then" (implication, →), "if and only if" (implication (equivalence), ↔); as well as "not" (negation, ¬); logical variables – propositional letters, propositional variables; and auxiliary signs – brackets. Sentences, formulas – variables and composite expressions formed from the variables combined with the connectors. In addition, a function in this context is the attribution of a value to each entity of a specific kind, in this case unary functions mapping formulas onto truth values. Finally, the validity of the composite expressions depends on the truth values of their constituent variables and the specific logical constants, i.e., truth-tables. Mendelson, *Introduction to Mathematical Logic*, 1-4.

[12] In predicate logic, there is a richer formal construct than in propositional logic; it consists of logical connectors: as in the former fn., and quantifiers: the universal quantifier ("for all") (∀) and the existential quantifier ("there exists") (∃), always combined with a variable, e.g., (∀x), (∃y); an infinite set of variables x, y, z , …; individual constants a, b, c, …; functional symbols of arity n (greater or equal to zero), which can take as argument any term (which is recursively constructed in terms of the constants, variables and functions); predicate constants or predicate letters P, Q, R, …; unary standing for the properties of entities P(a), and n-ary $Q_i(a_1,…,a_n)$ standing for the relations between entities, also considering the order between the logical constants. The well formed formulas are generated in a natural recursive way, starting with the atomic formulas, which are the syntactic descriptions that m terms fulfil a particular n-ary relation. Gamut, *Logic, Language and Meaning*, 66, 71. The satisfiability and validity of w.f. formulas are defined in a coherent way based on the (implicitly assumed) existence of a model for the primitive notions of set and a membership relation between them, satisfying, for example, the axioms of Zermelo-Fraenkel (ZF, sometimes with the axiom of choice, ZFC). S. Shapiro, 'Logical Consequence, Proof Theory, and Model Theory', in S. Shapiro (ed), *The Oxford Handbook of Philosophy of Mathematics and Logic* (Oxford 2005), 651-653. In this framework, one can express the most important formal setting of modern mathematics, for instance, the notions of sets, conjunction, disjunction, bi-conditional, element, principle of extensionality, subset, property, union, intersection, ordered pairs, (formal) geometrical and arithmetical notions and properties. Gamut, *Logic, Language and Meaning*, 83-87.

[13] Formal (mathematical) semantics, Mendelson, *Introduction to Mathematical Logic*, 49.

[14] Regarding the application of logical models in legal practice: syllogistic logic, J. S. Edwards, R. I. Akroyd, "Modelling rhetorical legal 'logic' – a double syllogism" (1999) 51 Int. J. Human-Computer Studies 1173; propositional, predicate and dialogical logic, E. T. Feteris, *Fundamentals of legal argumentation: a survey of theories on the justification of judicial decisions* (Dordrecht 1999), 29-31.

[15] The description of the construction, interpretation and systematisation of legal means within the application of an identified knowledge-theoretical construction and argumentative analytical model – i.e., the specific rationale, Latour, op. cit., 38, 54-55, 66.

The legal theoreticians of countries late to the table of the great civil law codification work of the 19th and early 20th centuries witnessed the extensive validation, mapping and prediction of natural phenomena due to refined methods and equipment; the conceptualisation of the link between the expanding synthesised natural empirical phenomena and the dynamically evolving expressive power of mathematics[21]; and the mathematisation of argumentative analysis[22], initially combined with a representational epistemological notion of supporting matter[23]. In a number of works by different authors considered to constitute a "period", the potential of different philosophical knowledge theoretical paradigms and models for argumentative analysis, as conditions for the legal referential activity, were investigated[24].

---

[16] The discursive application of mathematics to predetermined objects of mathematics, i.e., figures and numbers; and the intuitive dialectical gesture of making the axioms of mathematics intelligible, technè, dianoia, metaxu, geometry, arithmetic; dialectics, épistémè, nous, principle, idea – ideatum, Plato, *The Republic* vol. 2 (Cambridge Mass. 1969), 114-115. And a schemata for the analysis of formally valid premise-conclusion arguments, in which a conclusion is drawn from two premises, allowing a limited options of subject-predicate propositions (universal affirmative, universal negative, particular affirmative, particular negative). The terms acting as subjects and predicates refer to concepts, not to singular expressions, and the validity of inferences is based on quantifying expressions, such as *all* and *some*, and not on the conjunction *if (...then)* ("syllogistic logic"). Gamut, *Logic, Language and Meaning*, 9-11. Regarding the application of the identified rational subject in a legal context: "Die Römer /…/ suchen die Einheitsforderung, der die Griechen in der Anschaung des Seins genügen wollten, im Reich des Handelns zur Geltung zu bringen. Und dadurch werden sie zu den ersten Logikern des Rechts." E. Cassirer, "Axel Hägerström. Eine studie zur schwedischen Philosophie der Gegenwart', *Gesammelte Werke Hamburger Ausgabe*, band 21 (Hamburg 2005), 89. Regarding the critique of the ontological notion expressed in the systematisation of legal rules by application of a syllogistic argument schemata, see (A. Ross, 'Tû-Tû' (1957), 70 Harvard Law Review 812) – although there is no clear answer to the question of whether the terms in a syllogistic logic are things (non-empty classes) or linguistic expressions of these things.
[17] Regarding transcendental idealism, see n 1. Regarding the application of a transcendental idealism in a legal context: "Die reinen 'Verstandesbegriffe', die 'Kategorien', sind, nach Kant's Definition, nichts anderes als die Mittel, mit deren Hilfe wir 'Erscheinungen' /…/ buschstabieren, um sie als Erfahrung lesen zu können." /…/ "Die Rechtsbegriffen haben die gleiche Aufgabe der 'Synthesis' zu vollziehen, aber ihre Einheitsbildungen haben einen ganz anderen Charakter, da sie sich nicht auf einen Einheit von Wahrnehmungen, sondern von Handlungen beziehen.", Cassirer, op. cit., 92-93, 101.
[18] Regarding German idealism, see n. 2.
[19] Regarding legal constructions as the expression of the "Volksgeist", see "German Historical School of Jurisprudence", P. Becchi, "German Legal Science: The Crisis of Natural Law Theory, the Historicisms, and "Conceptual Jurisprudence"', in D. Canale, P. Grossi, H. Hoffmann (eds), *A Treatise of Legal Philosophy and General Jurisprudence; vol 9: A History of the Philosophy of Law in the Civil Law World, 1600-1900* (Dordrecht 2009) 202-203, 206-207.
[20] Regarding legal constructions as the expression of the "Geist", see "legal jurisprudence", P. Becchi, op. cit., 222-224.
[21] Modern philosophical idealism, E. Cassirer, *Substance and Function* (Chicago 1923); Cassirer, *The Philosophy of Symbolic Forms*.
[22] W. Demopoulos, P. Clark, "The Logicism of Frege, Dedekind, and Russel", 131-132.
[23] Regarding correlationist and representational, see n. 6. Logical positivism, C. Hempel, "Problems and Changes in the Empiricist Criterion of Meaning" (1950), 11 Rev. Intern. de Philosophie 41; and analytical philosophy, B.

## II. PRELIMINARY CONSIDERATIONS

### A. Natural Scientific Verification Criteria

The identification of phenomena by the application of approved natural scientific methods engendered a philosophically conceptualised distinction between verified and non-verified referential content[25]. Carried over to a legal context, the application of natural scientific truth criteria introduced the task of classifying traditional legal concepts as scientifically or unscientifically viable[26]. The instantiation of a physical existing world as an intersubjective validity implied the denial of the reliance on a rational subject or on an organisation of objects predetermined in a mathematical context as a cohesive and legitimising level of foundation for legal constructions[27].

### B. Non-correlationist Concept Theory

The philosophical project of the conceptualisation of the formal expressive character of mathematics in the natural sciences led to a replacement of the empirical and rational subject by a dynamically developing mathematical instance capable of projecting empirically verifiable relations and identities onto an obtuse materia, whether in the context of the natural sciences or with regard to the usage of symbolic forms in the context of the social sciences[28]. The

---

Russel, "On Denoting" (1905), 14 Mind, New Series, 479; P. F. Strawson, "On Referring" (1950), 59 Mind, New Serie, 320.

[24] The Scandinavian tradition, N. Kr. Sundby, "Legal Right in Scandinavian Analysis" (1968), 13 Natural Law Forum 72; J. Samuelsson, *Tolkning och utfyllning: Undersökningar kring ett förmögenhetsrättsligt tema* (Uppsala 2008), ch. 7. The difference between the Scandinavian and the North American tradition, E. Millard, "Scandinavian Realism, American Realism. An Essay of Characterisation." (2014), Revus: Journal for constitutional theory and philosophy of law 81.

[25] Ontological/scientific naturalism, n. 2; E. Cassirer, *Determinus und Indeterminus in der modernen Physik. Historische und systematische Studien zu Kausalproblem* (Göteborg 1937), 82. The link to methodological naturalism, W. V. Quine, *Ontological Relativity & Other Essays*, 83.

[26] A. Hägerström's "theory of correspondence", M. Lyles, *A call for scientific purity: Axel Hägerströms critique of Legal Science*, Rättshistoriskt Bibliotek, band 65 (Stockholm 2006), 232-235; T. Spaak, *A Critical Appraisal of Karl Olivecrona's Legal Philosophy* (Dordrecht 2014), 46-48.

[27] A. Hägerström's "value theory"/"value psychology", often referred to as value nihilism, P. Mindus, *A Real Mind. The Life and Work of Axel Hägerström* (Dordrecht 2009), 105-112.

[28] E. Cassirer developed a theory of concepts based on the replacement of the transcendental categorical level of synthetisation by a mathematical instance – characterised by its dynamic development, its relative, non-representational nature, supported by Einstein's theory of relativity, and the theoretical limit of its power of expressiveness, M. Friedman, "Ernst Cassirer and Contemporary Philosophy of Science" (2006), 10 Journal of the Theoretical Humanities 119. The theory, placing the natural sciences and the social sciences on the same level,

application of the theory of concepts in a legal context warranted the re-identification of legal concepts as non-correlationist symbolic forms[29] corresponding to the different distinguished uses of natural languages – the "expressive function", with an empirically verifiable meaning content deducible from the specific context, rendering a "substantial concept",[30] and the "representative function", with a meaning content deducible from the specific theoretical context, rendering a "functional concept"[31,32]. As the transcendental level of a rational subject was replaced by a dynamically evolving formal instance, the characterisation of the power of the expressiveness of formal mathematics as theoretically converging towards a limit was further introduced as a foundational basis of the legal systems[33].

*C. Mathematical Logical Model for Argumentative Analysis*

A subsequent attempt involved a turn towards the movement that, in the wake of Frege's "logico-mathematical inquiry into the concept of number"[34], investigated the inner logic or deep structure of natural languages through a comparison with the behaviour of mathematical number theory – initially paired with a representational epistemological notion and an informal inference rule[35]. Applied to legal theory, the instauration of the richer formal construct of predicate logic[36] warranted a re-identification of traditional legal concepts and the construction,

---

was labelled "modern philosophical idealism" and "philosophy of symbolic forms", Cassirer, *The Philosophy of Symbolic Forms*, 45-85.

[29] Concepts in natural languages given meaning through the relative, non-correlationist, meaning function (*Bedeutungsfunktion*), Cassirer, op. cit., 48.

[30] The expressive function of natural language (*Ausdrucksfunktion*) through a "substantial concept" (*Substanzbegriff*), Cassirer, op. cit., 52.

[31] The representative function (*Darstellungsfunktion*) through a "functional concept" (*Funktionsbegriff*), Cassirer, op. cit., 57.

[32] The legal concepts having a representative function were thus not considered as fixed to an idealistic or materialistic reference-referred amalgamate, but they were also not discarded as unscientific, U. Bindreiter, "The realist Hans Kelsen" in L. Duarte d'Almeida, J. Gardner, L. Green (eds.), *Kelsen revisited* (Oxford 2013), 117-118.

[33] Legal positivism, L. Green, "Positivism and Conventionalism" (1999), 12 Canadian Journal of Law & Jurisprudence 35. The relation between the theoretical limit of the expressiveness of mathematics, employed in Cassirer's theory of concepts, and Kelsen's *Grundnorm*, E. Winter, *Ethik und Rechtswissenschaft: Eine historisch-systematische Untersuchung zur Ethik-Konzeption des Marburger Neokantianismus im Werke der Hermann Cohens* (Berlin 1980), 191. Critique of the introduction of a contemporary idealistic notion as a foundation for the legal system, A. Ross, H. P. Olsen, "The 25th Anniversary of the Pure Theory of Law" (2011), 31 Oxford Journal of Legal Studies 243.

[34] G. Frege, *The Foundations of Arithmetic: A Logico-mathematical Inquiry into the Concept of Number* (2nd edn., New York 1960).

[35] See n. 4. Verificationism as an inference rule in a mathematical logical context, i.e., an informal semantics, n. 9, n. 10. An empirical epistemological notion in the context of mathematical logic, T. Uebel, "Logical Positivism-Logical Empiricism: What's in a Name?" (2013), 21 Perspectives of Science 58. Decline of logical positivism, R. Creath, "Every Dogma Has Its Day" (1991), 35 Erkenntnis 347.

[36] See n. 10.

systematisation and interpretation of legal propositions through the translation into a mathematical terminology of n-ary predicate symbols[37], constants and quantifiers[38].

The combination of empiric verification criteria, a representational epistemological notion and a formal mathematical ordering instance, distributing qualities and connecting individuals in different contexts, established the market, society and the state as a frame providing coherence between the different scripts and setting up a list of defined interests guiding the constructions and interpretations of legal propositions[39], leaving it to other disciplines, such as psychology, anthropology and sociology, to study the motivation of the individuals acting in and in relation to the legal scripts[40].

*D. The Non-correlational Epistemological Functional Concept vs. the Logical Relational Symbol*

The philosophical concept of symbols with an indicative, expressional and representational function and the logical mathematical n-ary predicate symbol share the endorsement of a formal instance regulating the distribution of relations between entities[41]. However, the relational logical symbol is a syntactic figure with the expressiveness of its set-theoretical provenience in analytical philosophy connected to an intuitive empirical epistemological and semantical notion[42]. The function symbol, on the other hand, is a philosophical knowledge theoretical concept proposing a non-correlational symbolic notion, conditioning all type of referential activities[43].

---

[37] The relational forms such as a R a (a bears the relation R to a; there are also ternary relations and so on) are acknowledged. In syllogistic and propositional logic, relations had not previously been considered as fundamental properties, for example, set-theoretical properties: sets, conjunction, disjunction, bi-conditional, element, principle of extensionality, subset, property, union, intersection, ordered pairs. W. Demopoulos, P. Clark, "The Logicism of Frege, Dedekind, and Russel", 131-133.

[38] The introduction of a formal syntax in the legal activity: the cumulative plurality of legal consequences; the disjunctive plurality of conditioning facts; semantic reference; the use of indicative relational symbols to express a transfer of rights. A. Ross, "Tû-Tû" 812; A. Wedberg, "Some problems in the logical analysis of legal science" (1951), 17 Theoria 246. Further regarding the relation between law and logic, J. Woleński, "Law and Logic in the 20th Century", in E. Pattaro, C.Roversi (eds.), *A Treatise of Legal Philosophy and General Jurisprudence; vol. 12:2: Legal Philosophy in the Twentieth Century: The Civil Law World* (Dordrecht 2016), ch. 26.

[39] The denial of the reliance on a rational knowing subject or on an intuitively accessible known object as a cohesive and legitimising level of foundation for legal constructions but the maintenance of the existing world as an intersubjective validity established quantifiable economic and inter-social factors as universally valid rectified interests, part of the weighing of values underlying legal constructions. – "Policy arguments in a rule based system", J. Bell, "Policy Arguments and Legal Reasoning", in Z. Bankowski, I. White, U. Hahn (eds.), *Informatics and the Foundations of Legal Reasoning* (Dordrecht 1995), 74-78. See the socio-anthropologic knowledge-theoretical terminology for the introduction of certain universally valid rectified interests, as opposed to the introduction of a fully transcendental foundational level: "mini-transcendence" and "metadispatcher" as opposed to "maxi-transcendence" and "dispatcher", Latour, *An Inquiry in to Modes of Existence*, 399-401.

[40] Epistemological verification criteria composed of the organised verified references from all types of scientific activities provide the individual with a verified world and make it possible to construct a knowledge schedule to predict the values guiding a certain individual in performing a certain task, for example, a judge. – Theory of "predictionism" for judges", A. Peczenik, *On Law and Reason* (Kluwer Academic Publishers 1989), 262-265. Social constructivism, J. Bell, op. cit. The adaptation of Ross' epistemological program in a Quinean replacement naturalism, J. v. Holderstein Holtermann, "Naturalizing Alf Ross's Legal Realism: A Philosophical Reconstruction" (2014), Revus: Journal for constitutional theory and philosophy of law 165.

[41] See n. 3 and n. 4.

*E. Continued Development of the Legal Rationale*

As we have seen, each combination of epistemological paradigm and argumentative analytical model, when entangled and applied as conditions for the legal referential activity, reveals its own rationality – the implication and potential therefore have to be assessed independently[44].

Granted the opportunity to look back at recent decades' rash development of information products, software programming, and algorithms as their controlling DNA, it was neither the introduction of natural scientific verification criteria nor a non-correlationist concept theory but the acknowledgement of the force in the formalisation of natural language with the help of logic – with an interest in the empiric dimension only as a semantical criterion of recognition[45] and, indirectly, as an epistemological notion[46] – that had the potential to change the process of constructing legal networks and the verification procedure for propositions within these networks[47].

Introducing a formal logic as the meta-language for legal activity implies, in addition to the challenge of mastering the technical difficulties, a potential change in the theoretical understanding of the linking of legal propositions with the non-legal content from different contexts, i.e., the creation of legal means – the identifying trait of legal activity.

---

[42] See n. 10 and n. 33. The Polish logician/mathematician A. Tarski introduced formal semantics in mathematical language, which was subsequently introduced for natural languages, and the consequent distinction between language viewed as an object of discussion, the object language, and language as the medium in which such a discussion takes place (i.e., the meta-language), A. Tarski, "The Semantic Conception of Truth and the Foundations of Semantics" (1944), 4 Philosophy and Phenomenological Research 341. The separation between object and meta-language also dissolved the problem posed in predicate logic, as developed by Frege, and later in analytical philosophy by, for example, B. Russel and P. F. Strawson regarding the relation between the signs and their references and the consequent distinction between denoting, referring and non-referring, and non-denoting signs, Gamut, *Logic, Language and Meaning*, 25.

[43] Regarding the belonging of the theoretical framework in Ross' article "Tû-Tû", the terminology and the examples used suggest a focus on the mathematical logical aspect of the construction, interpretation and systematisation of legal means, whereas the critique of the ontological aspect earlier in the legal theoretical environment applied representational epistemological notions, materialism and naturalism, suggesting a focus on the empiristic epistemological notion of logical postivism. Peczenik, *On Law and Reason*, 214.

[44] See n. 13.

[45] Mendelson, *Introduction to Mathematical Logic*, 49.

[46] Interest as empiric criteria of recognition in cognitive science and in neuro-psychology, N. Depraz, S. Gallagher, "Phenomenology and the Cognitive Sciences: Editorial introduction" (2002), 1 Phenomenology and the Cognitive Sciences 1.

[47] The development of artificial intelligence in legal practice, large-scale data analysis and data-centric legal systems, L. Karl Branting, "Data-centric and logic based models for automated legal problem solving" (2017), 25 Artificial Intelligence and Law 5. Critique of the period, A. Oskamp, M. Lauritsen, "AI in Law Practice So fat, not much" (2002), 4 Artificial Intelligence and Law 227.

In the pairing of a full formal logic with a non-empiric epistemological paradigm, the notion of the creation of a reference through ascending from a sensory particular[48], whether assisted by an object predetermined in a mathematical context[49], a knowing subject[50], or a mathematical instance[51], could be replaced by a procedure in which a speculative materia through a hiatus is attached to a reference, according to established methods in different types of referential activities[52], and linked to other references in accordance with the applied mathematical model[53] and with the application of empirical criteria of recognition[54]. In this way, the identifications are loosened from any intrinsic properties and allowed to play different roles in different scripts in an actor-network type of relationship[55].

It would thus be possible to combine a logical n-ary predicate symbol with a non-correlational epistemological notion to fix a one-to-one correspondence between typographical and semantic properties in a certain context[56].

---

[48] See n. 6. Also referred to as "quasi objects", Latour, *An Inquiry in to Modes of Existence*, 105-106, 426.
[49] n. 13.
[50] n. 1, n. 2.
[51] n. 24. Further, see the compatibility of the empirical subject and a naturalism paradigm – naturalising phenomenology, phenomenologising nature, J. Reynolds, "Phenomenology and naturalism: a hybrid and heretical proposal", 393.
[52] Regarding the notion of network and value: "/…/ under the term 'network' we must be careful not to confuse what circulates when everything is in place with the set up involving heterogenous elements that allow circulation to occur". Latour, *An Inquiry in to Modes of Existence*, 32. The characterisation of the legal referential activity: "To be sure, all the connected elements belong to different worlds, but the mode of connection, for its part, is completely specific /…/"., ibid., 38.
[53] n. 5.
[54] The dissolution of the cohesive and authoritative frame and the "essentialistic" notion of nature in correlational, representational, epistemological paradigms through an analysis of "category mistakes" – a mistake of direction as opposed to a mistake of the senses, Latour, *An Inquiry in to Modes of Existence*, 52-53. The instauration of an immanent non-representational reference-referred amalgamate and, hence, a non-correlational notion of nature, Papineau, "Against representationalism (about conscious sensory experience)", 324. The possible compatibility between an empiric subject and a non-correlational paradigm, Reynolds, op. cit., 393.
[55] The theoretical conditioning of different referential activities in a non-correlationalist, non-representational, epistemological paradigm – sociological relationalism, P. L. Berger, T. Luckmann, *The Social Construction of Reality* (New York 1966); Actor-Network-Theory, B. Latour, "On actor-network theory. A few clarifications plus more than a few complications." (1996), 47 Soziale Welt 369. The conditions for legal activity by the application of a formal logic paired with a non-empirical epistemological paradigm, Latour, *An Inquiry in to Modes of Existence*, ch. 13-14.
[56] In formal logic, this is achieved through the application of a formal structure, in the form of a chosen model, carrying the meaning of purely syntactic propositions, Shapiro, "Logical Consequence, Proof Theory, and Model Theory", 652.

Our aim is not to expand on this theoretical belonging but to highlight one aspect of the discussed legal "period", the acknowledgement of the extended formal construct of predicate logic, to create a link to the full formal logical context[57].

In this article, the first part serves as a demonstration of how a certain choice of meta-language conditions the legal referential activity. We introduce the creation, systematisation, and interpretation of legal means by using a formal syntax and distributive validation criteria. In the second part, we add a formal semantics, and in the third part, we observe whether the extended meta-language suits the legal activity through the formalisation of a specific legal statute regarding the purchase of immovable property[58]. Lastly, in the final part, we discuss the potential advantages that this new formal deductive framework[59] has for implementing initial artificial devices that can enhance the deductive abilities of legislators, researchers and practitioners of law.

### III. INTRODUCTION TO A RICHER FORMAL FRAMEWORK

*A. Introduction of Constants, Variables, Functions, Relations and Quantifiers*

The advantage of the extended logic vocabulary of predicate logic, as developed by Gottlob Frege and later by representatives of analytical philosophy such as Bertrand Russel and Peter F. Strawson, is that it offers a terminology for dealing with how objects are used as semantic support[60] for a textual formal context[61] and how certain kinds of such objects can be chosen to represent specific "qualities" of objects[62] and can be related to each other[63] to create textual/factual distinctions within a unified context[64].

---

[57] A logic combining a formal syntax and formal semantics, Shapiro, op. cit., 651-653.

[58] The following elements of the logic-based approach to the creation of a logical formalisation of a given legal text have been identified: "(1) legal texts have a determinate logical structure; (2) there is a logical formalism sufficiently expressive to support the full complexity of legal reasoning on these texts, and (3) the logical expression can achieve the same authoritative status as the text from which it is derived"., Branting, "Data-centric and logic based models for automated legal problem solving", 5.

[59] In D. Grossi, A. Rotolo, "Recent Developments in Legal Logic", in Enrico Pattaro, Corrado Roversi (eds), *A Treatise of Legal Philosophy and General Jurisprudence; vol 12:2: Legal Philosophy in the Twentieth Century: The Civil Law World* (Springer 2016), 743-751, are described new general and valuable results towards the logic formalisation of legal issues. However, they belong to a deontic-logic perspective, which can be viewed as an approach based on a modal logic perspective. Earlier attempts of a formalisation, based on Wesley Newcomb Hohfeld's ("Some fundamental Legal Conceptions as Applied in Judicial Reasoning" (1913), Yale Law Journal, 16; Fundamental Legal Conceptions as Applied in Judicial Reasoning" (1917), Yale Law Journal 710) and conceptions of rights and duties, include Stig Kanger, "Law and Logic" (1972), Theoria 38, 85; Lars Lindahl, *Position and Change – A Study in Law and Logic* (Dordrecht 1977); David Makinson, "On the formal representantions of rights relations: Remarks on the work of Stig Kanger and Lars Lindahl" (1986), Journal of Philosophical Logic 15(4), 403 and Marek Sergot, "Normative Positions", in X. Parent, R. van der Meyden, D. Gabbay, J. Horty and L. van der Torre (eds), *Handbook of Deontic Logic and Normative Systems* (College Publications 2013). Our approach, in contrast, starts with premises and principles coming from MFOL.

[60] n. 38.

[61] This notion can be understood as the model under consideration, i.e, the formal structure that carries the "meaning" of purely syntactic propositions. It is worth noticing here that the models themselves are formal structures and are usually based on the existence of a kind of "canonical" model, such as the existence of sets and a membership relation obeying the axioms of ZFC, Mendelson, *Introduction to Mathematical Logic*, 49.

Carried to the legal field, the objects/subjects (i.e., entities) that support the unified factual/textual-physical context of a legal system are persons, objects and the relations between them, taking place in specific spatial/temporal events[65]. The function of the legal system is to support certain types of interactions by lifting up certain objects/subjects (i.e., entities), letting them represent certain properties and relations (in a logical sense), gathering them under a "sign" (frequently referred to as a legal consequence or as a logical thesis within a formal implication) and relating them with each other through logical connectors such as implication, conjunction or equivalence in the form of "[conditions] imply [consequences]"[66]. The "systematisation" of a legal system consists of grouping the network of "logical sentences" related through logical connectors in the form of hypotheses imply theses[67].

The collection of specific objects/subjects (i.e., entities) that are "lifted up" to support a property or a relation[68] and to constitute a part of the factual/textual context of the legal system depends on how certain identified interactions, of which the property or relations are part, should be promoted[69].

In accordance with the formal logical vocabulary introduced above, a legal "expression", such as a "right"[70], can be employed either as a "property"[71] regarding a specific feature that a collection of objects possesses, in this case the type of use of an object, or as a reference to a specific relation[72] between several parties involving the achievement of such a property[73]. In the first case, the right is defined as the competences regarding the use of a specified object[74], supported expressly or implicitly by the specific legal system. In the second case, the specific relation and the underlying logical theses have to be identified before the validity of a statement containing the expression can be tried[75].

---

[62] Properties in a logical sense, n. 33.
[63] Relations in a logical sense, n. 33.
[64] The "model", constructed from the propositions in the object language, in our case the legal system.
[65] In informal predicate logic and, later, in analytical philosophy, the interpretation of signs after the translation into the logical language depends on the distinction between the character of the signs as denoting-/referring or non-denoting/non-referring, n. 38. In contrast, in classical first-order logic, the "models" once again turn out to be abstract entities such as natural, rational or real numbers, for example, n 56.
[66] Shapiro, "Logical Consequence, Proof Theory, and Model Theory", 652-654, 659, 663.
[67] See n. 13.
[68] In a logical sense.
[69] The legal script delegates different roles to identified actors, in the form of "[conditions] imply [consequences]", according to the legislator's considerations, Latour, *An Inquiry in to Modes of Existence*, ch. 13-14.
[70] We here chose as definiendum a claim equipped with a correlative duty, restricted to the domain of civil law. The conception comprises mainly property rights, including ownership and other limited rights. Wesley Newcomb Hohfeld, "Fundamental Legal Conceptions as Applied in Judicial Reasoning" (1917), Yale Law Journal 710-770, 717, 738.
[71] In a logical sense.
[72] In a logical sense.

We have so far introduced a part of the extended syntax, but on a semantic level, the empirical distributive validation procedure remains. What we want is the expansion of the argumentative analytic model with formal validation criteria, applying a formal structure, in the form of a chosen model, carrying the meaning of purely syntactic propositions[76].

*B. Many-sorted First-order Logic: Inspiration for an Extended Logical Formalism*

In classic many-sorted first-order logic (MFOL)[77], we consider several entities such as sorts, constants, variables, functions and relations structured by means of sound logical connectives and (universal and existential) quantifiers. Now, the models in this framework once again have a very abstract and linguistic nature since they emerge within the context of the standard framework of Zermelo-Fraenkel set theory with choice (ZFC)[78].

On the other hand, in our present case of finding syntactic-semantical formalisations of (purchase) statutes, we need to be able to speak of suitable frameworks having "models" that are located in the (physical) world, e.g., persons, properties, and purchase documents.

---

[73] The term could here refer to the cumulative "consequences" connected to the "condition" entering into a formally valid transfer agreement, that is, the reciprocal claims of delivery and payment, the competence to transfer, and the competence to demand the object from a third party holding the object – the so-called net effects, T. Håstad, "Derivative Acquisition of Ownership of Goods" (2009), 17 European Review of Private Law 725. The term could also refer to the cumulative or individual consequence and connected conditions related to the protection of certain identified parties, not being parties to the transfer agreement, C. v. Bar, E. Clive, H. Schulte-Nölke (eds.), *Principles, Definitions and Model Rules of European Private Law: Draft Common Frame of Reference (DCFR)*, vol. 5, (Munich 2009), sec. VIII.-1-202, p. 4263.
[74] Ross; "Tû-Tû", 819, 822. Hohfeld, "Some fundamental Legal Conceptions as Applied in Judicial Reasoning", "Fundamental Legal Conceptions as Applied in Judicial Reasoning", considered also the competences regarding the use of the object as a relational aspect, including the correlative duty. As referred to by Réka Markovich, "Understanding Hohfeld and Formalizing Legal Rights: The Hohfeldian Conceptions and Their Conditional Consequences" (2020), Studia Logica 108, 129, p. 132,: in later formalisations in deontic logic based on Hohfeld's theory, the directed aspected of the use has been abandoned by Stig Kanger, "Law and Logic" (1972) and Lars Lindahl, *Position and Change – A Study in Law and Logic,* but reintroduced by David Makinson, "On the formal represantations of rights relations: Remarks on the work of Stig Kanger and Lars Lindahl", and Henning Herrestad, Christen Krogh, "Obligations directed from bearers to counterparties" (1995), in *Proceedings of the International Conference on Artificial Intelligence and Law*, ACM, 210.
[75] Ross, "Tû-Tû", 819, 822.
[76] n. 56.
[77] Maria Manzano, *Extensions of first-order logic*, vol 19, (Cambridge 1996), ch. VI.
[78] Mendelson, *Introduction to Mathematical Logic*, 291-293.

Therefore, in our case, we aim to develop an extended logic framework for enhancing the semantical scope of these formal structures in regard to more pragmatic objects/subjects such as those considered in legal statutes.

## IV. FORMALISATION OF THE TRANSFER OF OWNERSHIP OF (IMMOVABLE) PROPERTY

Our next task is to employ the introduced logical framework to create a "model" for a chosen part of the unified factual/textual-physical context of the legal system. We have chosen the Swedish legal system, specifically the statute for the purchase of immovable property, as our object of formalisation[79]. One of the reasons is that the legal statute is constructed in accordance with a logical framework that acknowledges the relational form[80]. In addition, each of the notions involved in purchase laws such as (immovable) property, purchase, buyer, seller, purchase document and the obligations between the parties, together with their formal interrelations, seems to fit quite well with the deductive behaviour of an MFOL framework. Furthermore, the legal statute allows us to enhance the semantic scope of the formal structures by specifying the vocabulary and the attributed content of each employed variable in the vocabulary within the model.

### A. Towards a New Logic Formal Framework for Describing (Purchase) Statutes

In this section, in a new logic framework, we formalise the following sentence: "A purchase of real property is concluded through the drawing up of a document of purchase signed by the seller and buyer"[81]. Let us denote the previous sentence as SENT.

We take inspiration from the well-known MFOL framework[82]. Many aspects of this particular logic approach are suitable for our purposes because MFOL is based on a "multi-semantic" approximation; that is, the models for the theory are, by definition, divided into a collection of "sorts"[83] that can more easily model the diversity of entities that appear naturally in the syntactic description of (purchase) laws, e.g., real and movable properties, sellers, buyers and contracts. Furthermore, the temporal aspects are very important in our formalisation; therefore, we also use a temporal sort. In addition, all the sorts defined here have an implicit temporal component in the sense that they are considered to be in the present. For instance, if we say let A denote the sort representing the collection of all buildings in Stockholm, we mean by that the collection of all buildings in Stockholm in the present.

---

[79] The regulation of transfer of immovable property is provided in a unified regime, where the the identified relations constituting subdivisions are indicated in the heading of the different sections of the statute. "Purchase", "Rights and obligations of the seller and the buyer", "Priority on grounds of title registration", "Bona fide acquisition by virtue of title registration and the import of title registration in certain case.", Land Code *(Jordabalk (1970:974))*, [https://www.kth.se/polopoly_fs/1.476821!/Land_Code.pdf](https://www.kth.se/polopoly_fs/1.476821!/Land_Code.pdf)).
[80] n. 68, n. 70.
[81] ch. 4, sec. 1. Land Code. We here take the Scandinavian perception of the transfer of ownership as a starting point; see n. 70.
[82] n. 72.

Now, since the interpretations in our particular case are given by concrete entities existing in nature (such as people and properties), we call the logic emerging here "physical multi-sorted first-order logic" (PML).

Therefore, let us start fixing some initial terminology: Let $P_{nl}$ denote the collection of all natural or legal persons with a certain well-defined legal capacity[84], and let $Pr_R$ denote the collection of all real properties (in Sweden), i.e., well-defined pieces of land together with the fixed constructions inside of them[85]. By $Pr_M$, we mean the collection of movable properties, i.e., here, all entities in Sweden are virtually included except persons and real properties[86]; by Pr, we denote the union of $Pr_R$ and $Pr_M$, i.e., the collection of properties in Sweden; and by $w: P_{ph}(PrR\ )$, $w: P_{ph}(Pr_M)$ and $w: P_{ph}(\text{Pr})$, we mean that w is a finite collection of real properties, movable properties and both types of properties, respectively. Here, we use the functional symbol $P_{ph}(-)$ to indicate the similarity of this sort with the set-theoretic construction of the power set of a fixed set, i.e., the set whose elements are the subsets of the corresponding set.

In addition, we use the sub-index "ph" to note that we are dealing with a kind of "physical" and, simultaneously, formal construction very closed related to the idea of considering sub-collections of entities. Moreover, we need an additional relation $\in_{ph}: Pr \times P_{ph}(\text{Pr})$, used as follows: $s \in_{ph} w$ means that *the property "s" belongs to the collection of properties "w"*.

---

[83] For the present case, sorts can be understood as the conceptual unities that constitute the formal taxonomy where different "legal" actions take place.
[84] Code of Judicial Procedure *(Rättegångsbalk (1942:740))*, ch. 11, sec. 1.
[85] Land Code, *(Jordabalk (1970:974))*, ch. 1 and ch. 2. Identification of movable and immovable/real property, M. Lilja, "National Report on the Transfer of Movables in Sweden", in W. Faber, B. Lurger (eds.), vol. 5 *National Reports on the Transfer of Movables in Europe* (Munich 2011), 45-46.
[86] Lilja, op. cit., 45-46.

In addition, we use a sort $T$ to describe the temporal dimension of the events; for instance, an interpretation of this sort could be the classic mathematical notion of positive real numbers $\mathbb{R}^+$. We also use a binary relation to compare the moments when events happen, i.e., $\leq_{ph}: T \times T$. Finally, $D$ denotes the collection of contracts describing the purchase of one or more properties in Sweden.

Moreover, we now need to define the collection of relational symbols that we employ in our description. First, we denote the relation of purchase by $Pur: P_{nl} \times Pr \times P_{nl} \times T$. Therefore, the expression $(a, s, b, t) \in Pur$ would mean that *at time "t", person "b" has purchased property "s" from person "a"*.

Now, let us define the "purchase document" relation as follows: $PuDo: D \times P_{nl} \times P_{nl} \times P_{ph}(Pr) \times T$, and $(d, a, b, z, t) \in PuDo$ if and only if *"d" is the document of purchase of the properties in "z" by person "b" from person "a", signed by both of them at time "t"*.

Therefore, we can formalise our initial statement SENT with the notions defined above as follows:

$$(\forall a, b: P_{nl}) \left(\forall w: P_{ph}(Pr)\right) (\forall t: T)[(\exists v: Pr_R)\left(v \in_{ph} w\right)$$
$$\rightarrow \Big[(\forall s: Pr)(s \in w \rightarrow (a, s, b, t): Pur)$$
$$\leftrightarrow (\exists d: D)(\exists r: T)(\forall s: Pr)\left(s \in_{ph} w \rightarrow \left(r \leq_{ph} t\right) \wedge \left((d, a, b, s, r): PuDo\right)\right)\Big]$$

Let us explain in more detail the precise meaning of the previous sentence.

First, due to pragmatic reasons, we assume that SENT (implicitly) refers to all the possible persons who could be potential buyers and sellers in Sweden, who are essentially the collection of people in $P_{nl}$.

Second, SENT applies only for collections of properties that include at least one real property[87]. Thus, we included a sub-statement mentioning this fact explicitly, i.e., $\cdots (\forall t: T)[(\exists v: Pr_R)\left(v \in_{ph} w\right) \rightarrow \cdots$.

---

[87] "A purchase of real property is concluded through the drawing up of a document of purchase signed by the seller and buyer. The deed shall contain a statement of the purchase price and a declaration by the seller that the property is transferred to the buyer", ch. 4, sec. 1. Land Code.

Third, we assume that exactly at the time when the purchase document is signed by both parties, the purchase relation starts to be fulfilled between them, independently if at a future time the same properties will be purchased again.

Fourth, the central part of the sentence is just expressing the fact that $a$ has sold all the properties in $w$ (which includes at least one real property) to $b$ if and only if there is one purchase document that was signed by both parties at some time before (resp. immediately) when the purchase's relation is stated to be valid.

## V. TOWARDS AN ARTIFICIAL CO-CREATIVE LEGAL ASSISTANT

One of the advantages of using a kind of MFOL grounding framework to characterise legal statutes, i.e., propositions in a legal context, is that we can use very robust software such as the Heterogeneous Tool Set (HETS)[88] to specify many-sorted first-order concepts, such as the previous concepts, by means of the Common Algebraic Specification Language (CASL)[89]. Now, the HETS has many advantages for performing conceptual operations at a symbolic and semantic level. For instance, one can explicitly compute a formal conceptual blending of two (input) concepts by means of categorical colimits, where the commonalities can be manually codified throughout a generic space[90]. Furthermore, the HETS also has integrated consistency checkers for finding initial verifications/refutations of the (in-)consistency of the blended theories. This kind of tool can be very useful if one desires to compare logically similar (and very complex) laws from different countries (for example, within the European Union)[91] because one could codify them in a first-order logic language, compute the commonalities in terms of a generic space, perform the corresponding formal blend in the HETS and finally analyse the final blended space very carefully, not only from a human legal perspective (performed by researchers of law) but also with the powerful syntactic deductive tools of the HETS. In particular, this tool would considerably enhance the logic scope and accuracy of legislators, researchers and legal practitioners in their intellectual work.

---

[88] T. Mossakowski, C. Maeder, M.Codescu, *Hets User Guide -Version 0.99-* (Bremen 2013).
[89] M. Bidoit, P. D. Mosses, *CASL User Manual: Introduction to Using the Common Algebraic Specification Language* (Dordrecht 2004).
[90] G. Fauconnier, M. Turner, *The Way We Think: Conceptual Blending and the Mind's Hidden Complexities* (New York 2002).
[91] Examples of construction and systematisation of legal propositions in private law on a European level, *von Bar, Clive, Schulte-Nölke (eds.), Principles, Definitions and Model Rules of European Private Law: Draft Common Frame of Reference (DCFR)*; C. v. Bar, *Gemeineuropäisches Sachenrecht Band I, Grundlagen, Gegenstände, sachenrechtlichen Rechtsschutzes, Arten und Erscheinungsformen subjektiver Sachenrechte* (Munich 2015).

## VI. MAIN CONCLUSIONS

The main results of this work aim to open a very concrete new way of using the deductive power of classic formal logic frameworks (e.g., MFOL), together with the corresponding artificial realisations of (some aspects of) them (e.g., HETS), to improve, enhance, and facilitate research on and the creation of legal constructs at a global scale. In particular, one can perform a formalising procedure similar to that above not only for Swedish legislation regarding transfer of ownership but also potentially for any kind of legal statute by performing a suitable initial description of the sorts formalisation. Importantly, legislators, researchers and practitioners retain an essential role within our proposal since we aim to improve their deductive capabilities, not replace them.

Therefore, in conclusion, from this perspective, formal logic and law can work together in a new way, allowing us to construct artificial legal co-deductive assistants that can improve, for example, the (purely theoretical) deductive skills of theoreticians and practitioners of law.